\documentclass[10pt,twocolumn,letterpaper]{article}

\usepackage[accsupp]{axessibility}
\usepackage{wacv}              

\newcommand*{\rom}[1]{\expandafter\@slowromancap\romannumeral #1@}
\makeatother
\usepackage{graphicx}
\usepackage{amsmath}
\usepackage{amssymb}
\usepackage{booktabs}
\usepackage{times}
\usepackage{epsfig}
\usepackage{graphicx}
\usepackage{amsmath}
\usepackage{float}
\usepackage{amssymb}
\usepackage{svg}
\svgpath{{../imgs/}}
\usepackage{markdown}
\usepackage{multirow}
\usepackage{graphicx}
\usepackage{amsmath}
\usepackage{subcaption}
\usepackage{amssymb}
\usepackage{bbding}
\usepackage{booktabs}
\usepackage{markdown}
\usepackage{makecell}
\svgsetup{inkscapelatex=false}
\usepackage{amsmath}
\usepackage{algorithm}
\usepackage[noend]{algpseudocode}
\usepackage{mathtools,xparse}

\DeclarePairedDelimiter{\norm}{\lVert}{\rVert}
\usepackage{enumerate}
\usepackage[pagebackref,breaklinks,colorlinks]{hyperref}

\usepackage[capitalize]{cleveref}
\crefname{section}{Sec.}{Secs.}
\Crefname{section}{Section}{Sections}
\Crefname{table}{Table}{Tables}
\crefname{table}{Tab.}{Tabs.}

\def\wacvPaperID{1430} 
\def\confName{WACV}
\def\confYear{2024}

\begin{document}
\title{Self-Annotated 3D Geometric Learning for Smeared Points Removal}

\author{Miaowei Wang\\
University of Edinburgh\\
{\tt\small m.wang-123@sms.ed.ac.uk}
\and
Daniel Morris\\
Michigan State University\\
{\tt\small dmorris@msu.edu}
}
\maketitle

\begin{abstract}
There has been significant progress in improving the accuracy and quality of consumer-level dense depth sensors. Nevertheless, there remains a common depth pixel artifact which we call smeared points. These are points not on any 3D surface and typically occur as interpolations between foreground and background objects. As they cause fictitious surfaces, these points have the potential to harm applications dependent on the depth maps. Statistical outlier removal methods fare poorly in removing these points as they tend also to remove actual surface points. Trained network-based point removal faces difficulty in obtaining sufficient annotated data. To address this, we propose a fully self-annotated method to train a smeared point removal classifier. Our approach relies on gathering 3D geometric evidence from multiple perspectives to automatically detect and annotate smeared points and valid points. To validate the effectiveness of our method, we present a new benchmark dataset: the Real Azure-Kinect dataset. Experimental results and ablation studies show that our method outperforms traditional filters and other self-annotated methods. Our work is publicly available at \url{https://github.com/wangmiaowei/wacv2024_smearedremover.git}.

\end{abstract}

\section{Introduction}
\label{sec:intro}
\begin{figure}[t]
  \centering
    \includegraphics[width=1.0\linewidth]{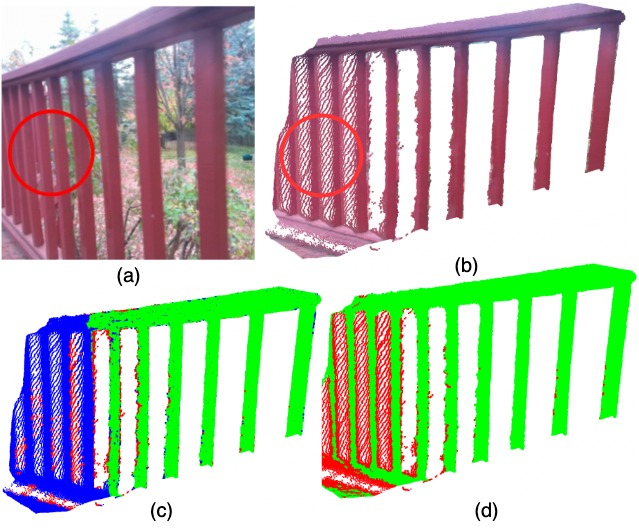}  
\caption{Example scene recorded by an Azure Kinect sensor with smeared points. The cropped color is shown in (a) and a colorized side view of the 3D point cloud is in (b).  Significant smearing can be seen between the vertical columns in the red circles.  In subplot (c), our method uses multiple viewpoints to automatically annotate smeared points (red) from valid points (green) and left uncertain points (blue).  Once trained, our method classifies pixels in a single frame as smeared or valid in subplot (d).}
\label{fig:onecols}
\end{figure}

While dense depth sensors have led to dramatic improvements in 3D computer vision tasks, including alignment~\cite{ICP_Alignment}, classification~\cite{Uy_2019_ICCV}, and reconstruction~\cite{KinectFusion}, they nevertheless still suffer from depth artifacts which can harm performance. Factors including scene complexity~\cite{RanSacForDetect}, hardware device conditions~\cite{Kinect2}, and sensor motion~\cite{MODKInectNoise} can adversely impact depth. Fortunately, consumer-level depth sensors have improved over the years~\cite{AKCOMpare}, with long-standing problems such as Gaussian noise, shot noise, and multi-path interference being alleviated. However, there continues to exist an important class of invalid depth points at the boundaries of objects, as shown in ~\cref{fig:onecols}. These points often interpolate between objects across depth discontinuities, and so we call them \textbf{smeared} points, in contrast to other outliers or random noise.  Our primary goal is to eliminate smeared points without harming other depth points, especially valid boundary details. 

A primary cause of smeared points is multi-path reflections.  Pixels on or adjacent to edge discontinuities can receive two or more infrared signal reflections; one from the foreground object and one from the background. Depending on the sensor circuitry, these multiple returns can result in a variety of artifacts and typically are interpolated between the foreground and background object. Common depth noise has a small bias compared to variance and low dependence on 3D shapes.  In contrast, smeared point noise, caused by multi-path interference, depends strongly on 3D scenes with one-sided distributions at object boundaries, see \cref{fig:onecols}.  These smeared points can be problematic for applications that use depth maps as they result in false surfaces in virtual worlds, blurring of fine 3D structures, and degraded alignments between point clouds.  These harms are compounded when multiple point clouds each having different artifacts are combined into an overall blurred point cloud.

\begin{figure}[H]
\centering
\includegraphics[width=1.\linewidth]{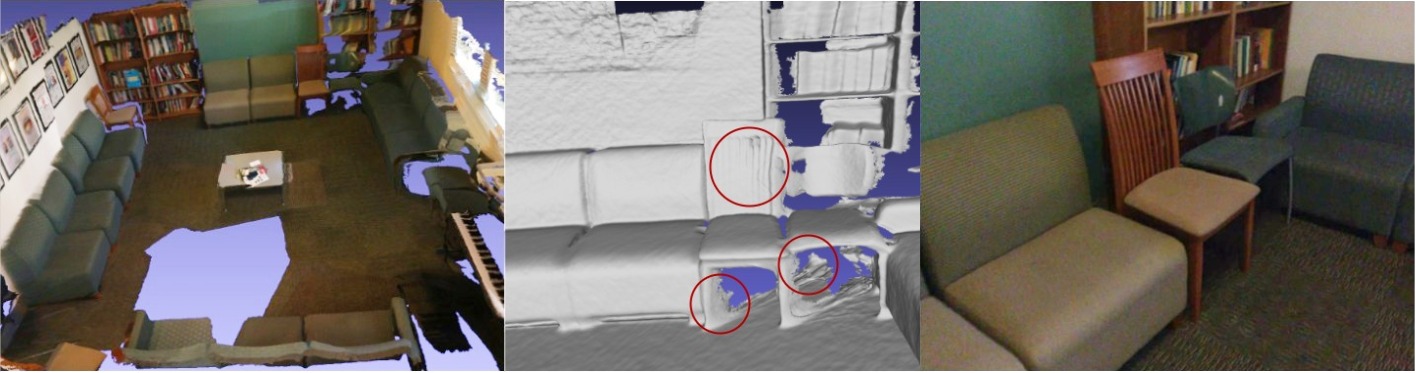}
\caption{A well-reconstructed 3D model example of ScanNet\cite{dai2017scannet} (left) contains smeared points in the red circles (middle), and the color image (right) is provided for comparison.}
\label{fig:scannet}
\end{figure}

Now, improvements in sensor processing have given modern sensors the ability to remove some of these smeared points, particularly when there is a large gap between the foreground and background objects. Nevertheless, smearing across smaller depth discontinuities is not solved due to the difficulty in distinguishing occlusion effects from complex shape effects, and as a consequence smeared points continue to plague the otherwise high-quality depth images, shown in \cref{fig:onecols}.  A variety of hand-crafted filters~\cite{MedianFilter,gaussian_filter, Bilateral} can be used to reduce noise in-depth maps, but we find that they perform poorly in removing smeared points or else result in overly smoothed surfaces.  A data-driven approach would be preferable, but these face the difficulty of acquiring sufficient ground truth which is expensive and time-consuming to obtain. More importantly, it should be pointed out that smeared points extensively exist in current famous RGB-D datasets such as LaMAR\cite{sarlin2022lamar}, NYU Depth V2\cite{Silberman:ECCV12}, and ScanNet\cite{dai2017scannet}. Thus the smeared point is not a niche problem. And there are still smeared points in their provided well-reconstructed ground truth 3D models shown in \cref{fig:scannet}, which prevents getting clean depth maps from large-scale off-the-shelf datasets.  
\vspace{-3.5mm}
\vspace{-4.5mm}Another approach is to create synthetic datasets~\cite{10.1145/3130800.3130884} with known ground truth, but these are limited by how well they model both the sensing and the environment.  Unsupervised domain adaption~\cite{RADU, Agresti_2019_CVPR} can address this to some extent.  However, approaches using multiple different frequencies~\cite{DeepMultis} from the same position, or using multiple cameras~\cite{Self-supervised} create significant overhead in acquisition. 

The goal of this paper is to overcome the difficulty in acquiring ground truth data for hand-held depth sensors by developing a novel self-annotated method for eliminating smeared points.  This avoids the need for building complex optical sensing models, and it also eliminates the need for expensive manual annotations of data.  Instead, our approach leverages the dense depth sensing capabilities of these sensors, along with a multi-view consistency model to automatically self-annotate points.  In this way, data can be rapidly acquired without human annotation and used to train a smeared-point remover.

In order to evaluate this method, fifty different real scenes both indoors and outdoors have been collected. Comprehensive experiments on these datasets and ablation studies further demonstrate the core idea in this paper that multi-frame self-annotation can effectively train a smeared point remover.    In summary, our contributions are:
\vspace{-0.5em}
\begin{itemize}
\item To our knowledge, we propose the first self-annotation technique for smeared points detection that applies geometric consistency across multiple frames.
\vspace{-0.5em}
\item By combining self-annotated labels with a pixel-level discriminator, we create a self-annotated smeared point detector.
\vspace{-0.5em}
\item We introduce a new real smeared points dataset (AzureKinect) using the Azure Kinect sensor as a benchmark.
\vspace{-0.5em}
\item We validate our design choices with several ablations.
\end{itemize}

\begin{figure*}[t]
  \centering
   \includegraphics[width=1\linewidth]{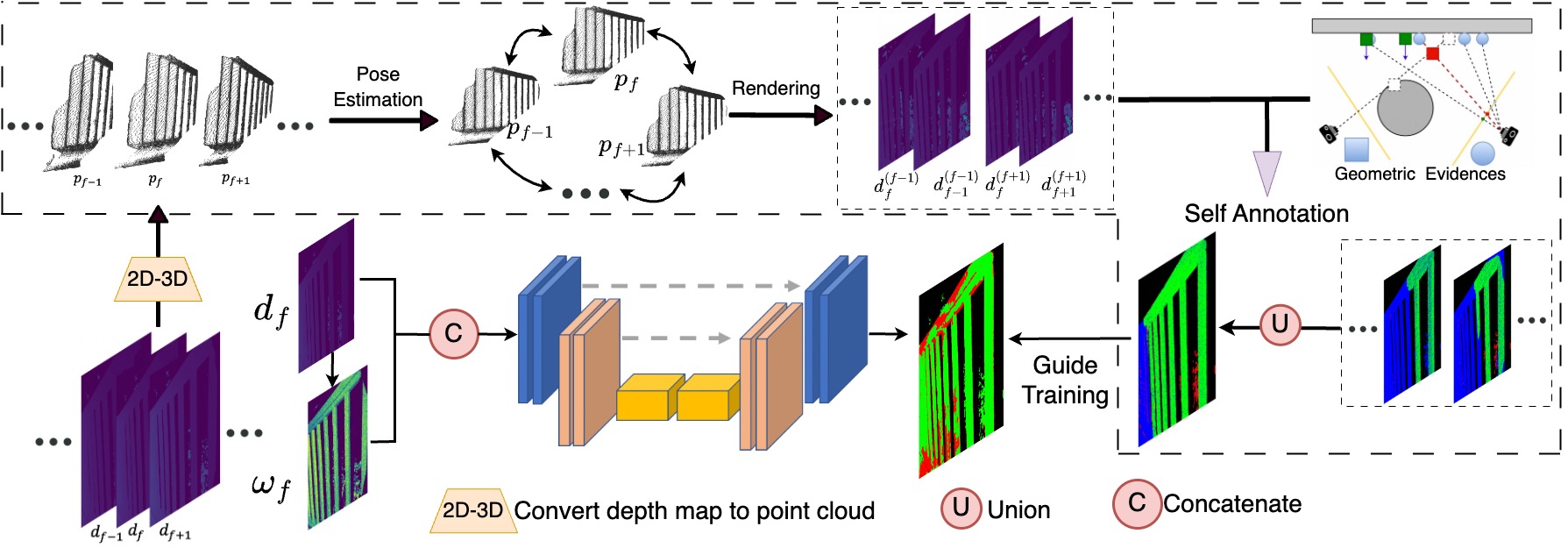}  
  \caption{Our self-annotated architecture for smeared point removal.  Training scenes are recorded with a hand-held sensor.  Multi-frame pose estimation aligns these frames.  Then geometric consistency is used to annotate smeared(red), and valid(green) pixels for all frames with left points as unknown(black).  Then a U-Net-based classifier is trained to identify smeared points in each frame. }
  \label{fig:diagram}
\end{figure*}
\section{Related Work}
\label{sec:relatedwork}
Obtaining noise-free, dense depth from raw, low-quality measurements has received significant attention. Before the rise of data-driven techniques, especially deep learning, numerous hand-crafted filters were designed to alleviate noise by referencing neighboring pixels, such as median filter~\cite{MedianFilter}, Gaussian filter~\cite{gaussian_filter}, Bilateral filter~\cite{Bilateral}, etc.  Early work to remove outliers introduced density-based and statistical methods~\cite{DensityOutlier,DensityBased,DcborDens}, while geometric and photometric consistency between depth maps and color images~\cite{Consistency,Kolev_2014_CVPR} was also used to detect outliers. As for time-of-flight multipath interference (MPI), multiple different modulation frequency measurements~\cite{Bhandari:14,bhandari2014resolving} of the same scene are collected to improve depth quality.  In contrast to these methods requiring multiple measurements at different frequencies, our method requires only a single-frequency depth map. 

Even before deep learning techniques were widely adopted, convolution and deconvolution techniques~\cite{Deconv} were proposed to recover time profiles only using one modulation frequency. DeepToF~\cite{10.1145/3130800.3130884} uses an autoencoder to correct measurements based on the observation that image space can provide most of the sources for MPI. Continuing the classical multi-frequency method, a multi-frequency ToF camera~\cite{DeepMultis} is integrated into the network design to preserve small details based on two sub-networks. RADU~\cite{RADU} updates depth values iteratively along the camera rays by projecting depth pixels into a latent 3D space. These supervised learning methods heavily rely on synthetic datasets generated by a physically-based, time-resolved renderer~\cite{JaraboSIGA14} that uses bidirectional ray tracing which is much more time-consuming to render one realistic depth map. To shrink the gap between real and synthetic datasets, DeepToF~\cite{10.1145/3130800.3130884} learns real data's statistical knowledge by auto-learning encoders while RADU~\cite{RADU} applies unsupervised domain adaptation by investigating a cyclic self-training procedure derived from existing self-training methods for other tasks~\cite{pmlr-v119-kumar20c, Lee2013PseudoLabelT, Self_Training}.  Additionally, an adversarial network framework can be used to perform unsupervised domain adaptation from synthetic to real data~\cite{adversarial_unsuper}. All these methods depend on the reliability of the simulated dataset. Moreover, current self-supervised methods either require a setup of multiple sensors placed in precomputed different positions based on photometric consistency and geometric priors~\cite{Self-supervised} or build noise models by assuming noises follow some random distribution around normal points~\cite{TotalDenoise,9812392} which leads to low availability when processing real scenes. In contrast to these approaches, our method operates in a self-annotated manner directly on real scene data without relying on complex scene formation models or specific noise models, or synthetic datasets.

\section{Method}

\subsection{Approach Summary}

This paper divides the smeared point removal into two distinct components: (1) a pixel annotator and (2) a pixel classifier, which are illustrated in \cref{fig:diagram}. Advances in correcting depth offsets~\cite{TotalDenoise,Self-supervised,RADU,10.1145/3130800.3130884} lead to high-quality depth estimates for the majority of depth pixels, leaving a typically small fraction of invalid or smeared pixels.  With these pixels often having large errors, our approach is to identify them for removal rather than correct their depth. Thus smeared point removal is a classic semantic segmentation problem and if we had sufficient annotated data, then a supervised classifier could be trained to perform this task. The challenge is how to obtain sufficient annotated data, as manual annotation is time-intensive and expensive.

In this section first, we describe two types of evidence for classifying pixels as either smeared or valid. By accumulating this evidence from multiple scene views, we create an automated smeared-pixel and valid-pixel annotation method. We then use these annotations to train a supervised single-frame smeared pixel classifier 

\subsection{Multi-View Annotation}


Typically smeared pixels occur between objects along rays that graze the foreground object. Now, as the viewpoint changes, these grazing rays change orientation and the resulting location of any interpolated points along these rays will also change. On the other hand, 3D points on objects will remain consistent, or at least overlap, between differing viewpoints. Thus we conclude if a pixel has been observed from multiple viewpoints with differing rays, the pixel must be a valid surface pixel and not a smeared point.

\begin{figure}
\centering
\begin{subfigure}{.4\textwidth}
  \centering
  \includegraphics[width=1\linewidth]{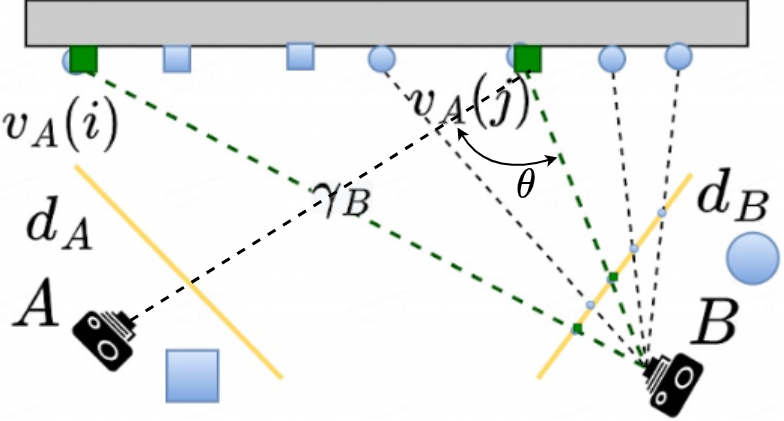}  
  \caption{Multi-Viewpoints}
  \label{fig:sub-second}
\end{subfigure}
\vskip\baselineskip
\hspace{-0.8cm}
\centering
\begin{subfigure}{.26\textwidth}
  \includegraphics[width=1\linewidth]{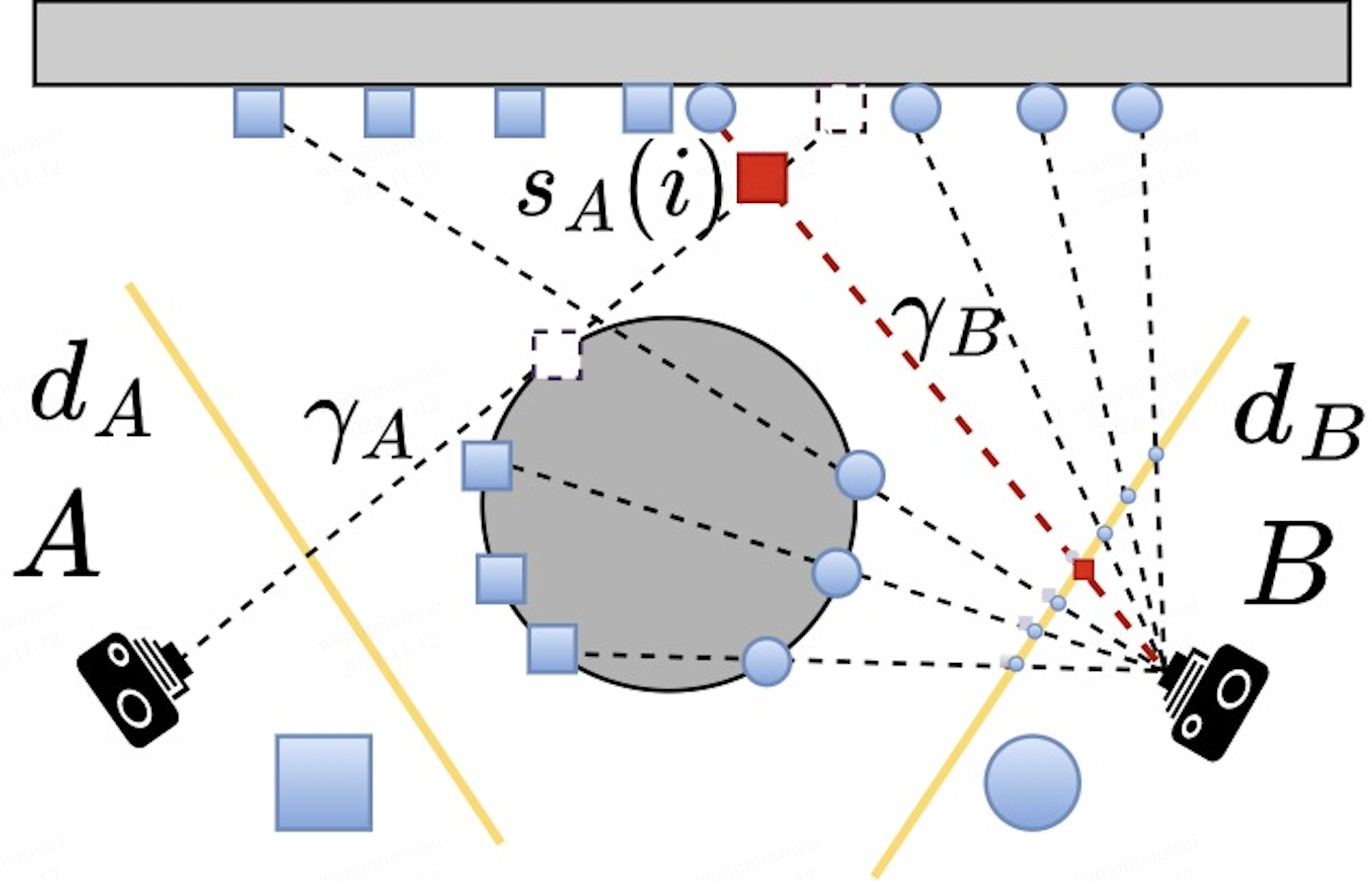} 
  \caption{See-Through Behind}
  \label{fig:sub-third}
\end{subfigure}
\hfill
\begin{subfigure}{0.25\textwidth}
  \hspace{+3.3cm}
  \includegraphics[width=1.0\linewidth]{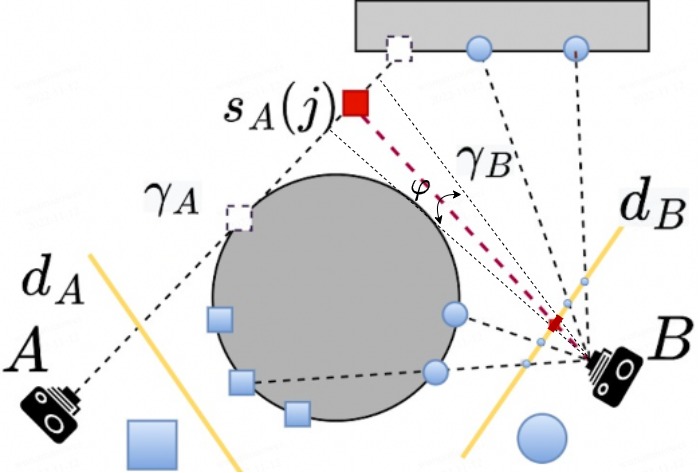}
  \caption{See-Through Empty}
  \label{fig:sub-fourth}
\end{subfigure}
\caption{Geometric evidence used for annotating our depth maps. Multi-Viewpoints evidence for valid points (green) is shown in the top row. Two cases of See-Through evidence for smeared points (red) are shown in the bottom two rows.}
\label{fig:evidence}
\end{figure}

An example of \textbf{multi-viewpoint evidence} is shown in~\cref{fig:sub-second}.  Points $v_A(i)$ and $v_A(j)$ are observed from separate viewpoints $A$ and $B$ and thus determined to be valid points. Now if the distance between viewpoints is small or the distance to the pixels is large, smeared pixels can coincide spatially. To avoid this, we use the angle $\theta$, always less than 90\textdegree, between the viewing rays of coincident points as a confidence measure in a point that is valid, and the confidence score $\emph{c}$ can be modeled as \cref{equ:confidence_score}
\begin{equation}
\emph{c} = \sin^2{(\theta)} 
\label{equ:confidence_score}
\end{equation}
The normalization is applied to the confidence score $\emph{c}$ to be in the range between 0 and 1. \cref{table:realsmeared} validates this design. 
\subsection{Space Carving Annotation}

The second category of evidence we gather has to do with space carving.  Smeared points, by definition, float off the surface of objects.  Now if a ray measuring a depth pixel passes through the location of a 3D point, then this is evidence that that pixel is not actually at that location but is most likely a smeared pixel.  

We divide {\bf see-through evidence} for smeared points into a case of positive evidence(See-through Behind) in ~\cref{fig:sub-third} and negative evidence(See-through Empty) in ~\cref{fig:sub-fourth}.  In both cases, a point is concluded to be a smeared point if another viewpoint can see through it.  In the first case, \cref{fig:sub-third}, a ray $\gamma_B$ from the camera at location B passes through a point $s_A(i)$, observed from location A, and measures a point behind $s_A(i)$, from which we conclude $s_A(i)$ is a smeared point.  In the second case \cref{fig:sub-fourth}, a point $s_A(j)$ observed by $A$ should be visible to viewpoint $B$, and yet there is no measurement along the ray $\gamma_B$, either closer or farther than $s_A(j)$.  To conclude from this negative evidence that $s_A(j)$ is a smeared point we expand the ray $\gamma_B$ between the sensor and $s_A(j)$ to a conical section with angle $\varphi$ and require no points are observed from $B$ within this, which eliminates the case of grazing rays being blocked and incorrectly inferring a smeared point behind them. The conical section angle $\varphi$ is a regularization term in See-Through Empty and larger values mean fewer detected smeared points with higher confidence. A naive quick equivalent implementation of $\varphi$ is applying a sliding window in the depth map. No reference points around the detected smeared point in a larger window size mean higher $\varphi$.       In our experiment, the sliding window with size $3\times3$ is used to filter unconfident self-annotated smeared labels in See-through Empty.

\subsection{Geometric Label Generation}

Automated pixel annotation involves combining the geometric evidence for valid and smeared points to a sequence of depth images. We note that pixels for which none of the two pieces of evidence apply will have an unknown categorization. To convert geometric evidence among multiple frames to geometric labels trained for the network, we assume that a depth sensor is moved around a rigid scene, typically by hand, and gathers depth frames $\{d_{f-m//2},\cdots,d_{f+m//2}\}$ from totally $m+1$ consecutive viewpoints, and from which 3D point clouds $\{p_{f-m//2},\cdots,p_{f+m//2}\}$  are created. Then the first step is to align all viewpoints, which is achieved by multi-frame Iterative Closest Point (ICP)~\cite{PoseGraph}.  The result of this alignment is an array of sensor viewpoints and a single-point cloud with each point having a viewing ray to the sensor from which it was gathered.  To determine the point visibility we use ray-tracing through rendering as described next.

{\bf Pixel Rendering} Applying our geometric evidence requires visibility reasoning for all pixels, which is performed using rendering.  We denote a pixel observed in frame $f$ as $p_f$ with coordinates $(u_f,v_f)$ and depth $d_f$.  Since we know all camera poses, the pixel can be projected into any other frame $f'$, represented as $p_f^{(f')}$ with coordinates $(u_f^{(f')},v_f^{(f')})$ and depth $d_f^{(f')}$.  This defines a mapping from original pixel coordinates to coordinates in any other camera:
\begin{equation}
I: (u_f,v_f) \rightarrow (u_f^{(f')},v_f^{(f')})
\label{equ:index_map}
\end{equation}

Additionally, due to different parameter settings and depth-buffering mechanisms between our renderer and the actual depth sensor, point cloud $p_{f'}$ should also be reprojected to the depth map $d_{f'}^{(f')}$ with the same renderer of $d_f^{(f')}$ when applying our geometry evidence. 
\begin{table}
\centering
\scalebox{0.70}{
\centering
\begin{tabular}{ccccc}
   \toprule
   Multi-Viewpoints & See-Through Behind&See-Through Empty & Inference \\
   $v_f=1$ & $b_f=1$ & $e_f=1$ & \\
   \midrule
   \Checkmark &$-$&$-$& Valid \\
   $-$&\Checkmark&$-$& Smeared \\
   $-$&$-$& \Checkmark & Smeared \\
   $-$&$-$& $-$& Unknown \\
   \bottomrule
\end{tabular}}\\
\caption{Find valid and smeared points using the multi-view consistency and ray-tracing model respectively.}
\label{Evidence}
\end{table}

The geometric evidence can be gathered into three binary variables for each pixel $\{v_f,b_f,e_f\}$ with each taking values $[0,1]$.  Here $v_f=1$ indicates valid pixel evidence as it is viewed in multiple frames as in~\cref{fig:sub-second}, while $b_f=1$ indicates smeared pixel evidence due to See-Through Behind in~\cref{fig:sub-third}, and $e_f=1$ indicates smeared pixel evidence due to See-Through Empty as in~\cref{fig:sub-fourth}.  These are summarized in~\cref{Evidence}.  Then, our algorithm to use this evidence to label pixels is shown in~\cref{alg:euclid}. 
\begin{algorithm}
\caption{Algorithm to automatically generate geometric labels for each pixel. Small constants $\epsilon$ and $\delta$ are set according to pixel depth noise. }
\label{alg:euclid}
\begin{algorithmic}[1]
\State For target frame $f$, initialize: $b_{f},e_{f},v_{f}=0$
\For{each $(u_f,v_f)$ in $d_f$}\
\For{each $f' \in [f-m//2,f-1]$ and $[f+1,f+m//2]$}
     \State Rendering new maps $d_f^{(f')}$ and $d_{f'}^{(f')}$
     \State Index buffer $I: (u_f,v_f)\rightarrow (u_f^{(f')},v_f^{(f')})$
       \If{$(u_f^{(f')},v_f^{(f')})$ is inside frame $f'$}
       \State $k=d_f^{(f')}-d_{f'}^{f'}$
                  \If{$|k|<\epsilon$}
                  \State $v_{f}(u_f,v_f)=1$ \Comment{Valid pixel} 
                  \ElsIf{$k<-\delta$}
                  \State $b_{f}(u_f,v_f)=1$ \Comment{See-Through Behind}
                  \ElsIf{$k=d_f^{(f')}$}
                  \State $e_{f}(u_f,v_f)=1$ \Comment{See-Through Empty}
                  \Else
                  \State continue;          \Comment{Unknown category}
                  \EndIf
            \EndIf
        \EndFor
    \EndFor
\end{algorithmic}
\end{algorithm}

In this algorithm, pixels observed in a target frame $f$ are labeled as valid or smeared by doing a pairwise comparison of rendered depths, $(d_{f}^{(f')}, d_{f'}^{(f')})$, in each of the other reference frames, $f'$.   The number of used reference frames per sequence, $m$, can be varied, although here we used $\textbf{m=4}$ which enabled good multi-frame alignment.  

\textbf{Why train a model} rather than directly applying such heuristics? We note that while a multi-frame annotation can be used on its own to remove smeared points, it leaves a significant fraction of points unlabeled ($\textbf{85}\%$ in our AzureKinect training sets).  Relying on this also requires static frames and camera motion, and creates latency.  Thus, we use the annotation to train a single-frame network to do the eventual smeared point detection.

\begin{figure}
\centering
    \includegraphics[width=1.0\linewidth]{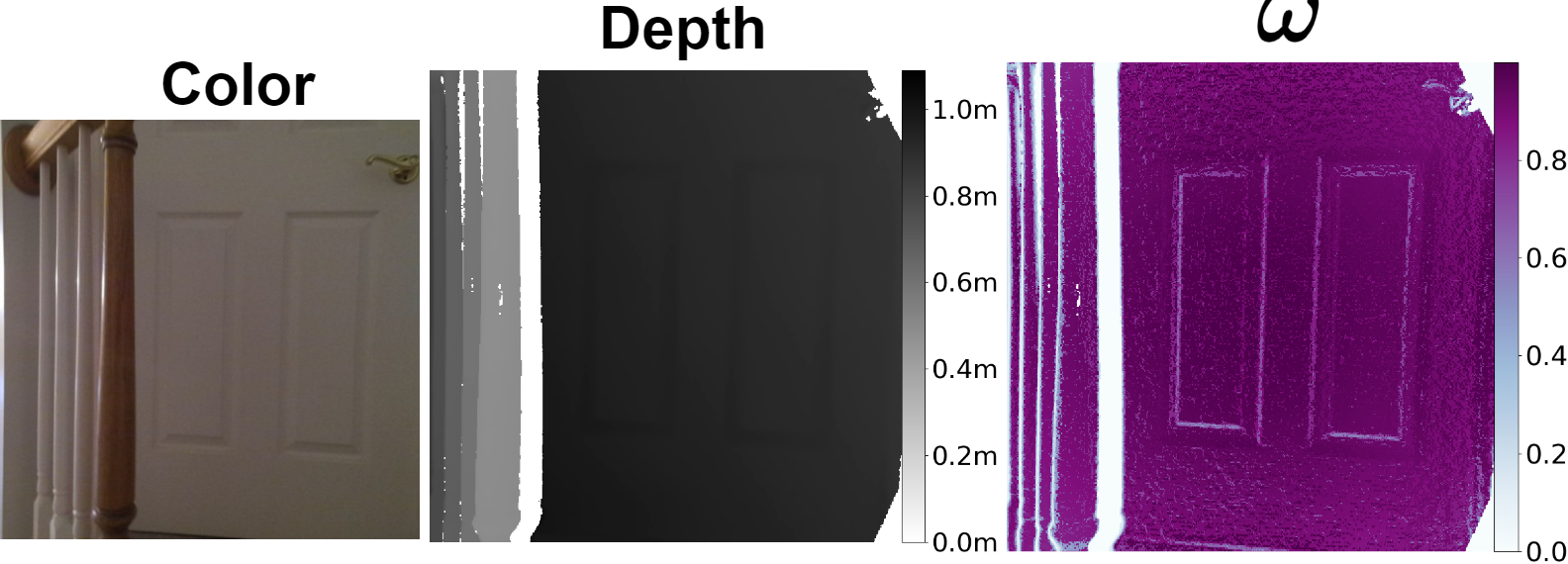}
\caption{Visualization of the normal view on an indoor scene and values of the boundary are lower compared to non-boundary areas. NOTE: Missing values in the corners of the depth map are directly related to the field-of-view(FoV)\cite{kurillo2022evaluating} of the depth camera.}
\label{fig:normal-view}
\end{figure}

\subsection{Depth Normals}

We anticipate that surface normals will provide useful cues to pixel classification. In particular, smeared pixel normals are often orthogonal to the viewing ray.  
Surface normals can be computed efficiently and directly from depth maps~\cite{7335535}.  We will specify the normal vector $n(u,v)$ at a pixel location $(u,v)$ in the depth map $d$.  This normal can be specified as the perpendicular to a facet connecting the 3D pixel $p(u,v)$, and its neighbor pixel location. Following~\cite{7785084}, we define $\omega(u,v)$ to be the inner product of the viewing ray unit vector and the normal unit vector:
\begin{equation}
    \omega(u,v) =n(u,v)^T \frac{p(u,v)}{||p(u,v||}
\label{equ:omega_atr}
\end{equation}
As shown in \cref{fig:normal-view},  an $\omega$ of 1 indicates a surface perpendicular to the viewing ray,  while an $\omega$ of 0 indicates an orthogonal surface.

\subsection{Smeared Classifier and Loss Function}
Some off-the-shelf 2D-based segmentation network is adapted here as our smeared classifier rather than a 3D segmentation backbone for three considerations: (1) it is lightweight and fast, (2) depth maps are directly obtained by the sensor when processing raw IR map, and (3) the smeared points generally deviate along the viewing ray, i.e. z-axis which indicates using a z-buffer is sufficient.  Our smeared classifier $\Psi$ maps an input $\phi=\left\{d,\omega \right\}$ consisting of a depth map and corresponding ray inner products, to an output consisting of the smeared probability \emph{p} as:
\begin{equation} \Psi: \phi \rightarrow \emph{p} \end{equation} 
We use a binary cross-entropy loss function with the above self-generated geometric labels:
\begin{equation}
    CE = -(b+e)\cdot \log{\emph{p}}-v\cdot \log(1-\emph{p})
\label{eq:cross_entropy}
\end{equation}
To balance both smeared and valid points, weights based on geometric label results are used here as  \cref{eq:weights}
\begin{equation}
w_\emph{k} = 1- \frac{\norm{\emph{k}}_{0}}{\norm{v}_{0}+\norm{b}_{0}+\norm{e}_{0}},\emph{k}\in \{b,e,v\}
\label{eq:weights}
\end{equation}
Besides, the confidence score $\emph{c}$ for the valid label is also considered to improve robustness as \cref{eq:finalloss}
\begin{equation}
L = -\alpha \cdot (w_{b}b + w_e e)\log{\emph{p}}- \beta \cdot \emph{c} w_v v\log{(1-\emph{p})}
\label{eq:finalloss}
\end{equation}
In the above final loss equation \cref{eq:finalloss}, $\alpha$ and $\beta$ are two hyper-parameters for fine-tuning in experiment sections.
\section{AzureKinect Dataset}
To validate the effectiveness of our methods, the real scene datasets using Azure Kinect were collected: we captured a total of 50 indoor and outdoor scenes using the Azure Kinect sensor, one of the state-of-the-art consumer-level cameras in the market. For each scene, we shoot 5 to 10 seconds with the hand-held camera moving without any speed or direction restraint under 5HZ operation frequency. And then a total of 1936 pairs of depth and color frames of real scenes are captured. Like some published datasets such as NYU Depth V2\cite{Silberman:ECCV12}, AVD\cite{active-vision-dataset2017}, GMU Kitchen\cite{georgakis2016multiview}, etc, our dataset provides pairs of color and depth information sharing the same resolution($1920\times1080$), as shown in \cref{table: properties}, by transforming depth image to the color camera and doesn’t hurt raw frame contents. And we also provide raw depth maps with resolution $640\times576$. Since there are currently no depth sensors on the market that can effectively avoid smeared points, we resort to manually annotating 11 typical frames for 11 different scenes respectively to get ground truth.  To ensure the accuracy of the annotation, human annotators are required to carefully observe the whole video clip for each test scene and modify GT labels several times repeatedly, which results in a single depth frame costing a human annotator about 6 hours. To our knowledge, our AzureKinect dataset exceeds existing published real ToF datasets in both size and resolution, see \cref{table: properties}, and is the only dataset provided with pose information for different views of the same scene. Therefore, our dataset lays a good foundation for future work on this new problem though the test set is admittedly small in size. 
\begin{table}
\centering
\scalebox{0.70}{
\begin{tabular}{cccccc}
   \toprule
   Dataset&Type&GT&Size&Resolution&Pose \\
   \midrule
   S1 \cite{DeepMultis}&Syn&Yes&54&$320\times240$&No \\
   S2 \cite{Agresti_2019_CVPR}&Real&No&96&$320\times239$&No \\
   S3 \cite{Agresti_2019_CVPR}&Real&Yes&8&$320\times239$&No \\
   S4 \cite{Agresti_2019_CVPR}&Real&Yes&8&$320\times239$&No \\
   S5 \cite{DeepMultis}&Real&Yes&8&$320\times239$&No \\
   \cline{1-6}\\
   FLAT\cite{guo2018tackling}&Syn&Yes&1200&$424\times512$&No \\
   \cline{1-6}\\
   Cornell-Box\cite{RADU}&Syn&Yes&21300&$600\times600$&No \\
   \cline{1-6}\\
   Zaragoza\cite{10.1145/3130800.3130884}&Syn&Yes&1050&$256\times256$&No\\
   \cline{1-6}\\
   \textbf{AzureKinect}&Real&No&1920&$1920\times1080^*$&Yes\\
   \textbf{AzureKinect(GT)}&Real&Yes&11&$1920\times1080^*$&No\\
   \bottomrule
\end{tabular}}
\caption{Properties comparisons of related datasets. GT refers to Ground Truth, while the size is the total number of frames. *AzureKinect dataset provides pairs of color and depth maps sharing the same resolution $1920\times1080$, also with raw depth map ($640\times576$ resolution) provided.} 
\label{table: properties}
\end{table}

\section{Experiments}
Deep learning models from similar tasks: multi-path interference removal (DeepToF~\cite{10.1145/3130800.3130884}), image semantic segmentation (UNet~\cite{Unet},  DeepLabV3+~\cite{Deeplabv3plus},  Segformer~\cite{xie2021segformer}), are used as the removal backbones based on our self-annotated framework. The self-annotated method DeepDD~\cite{Self-supervised} for removing regular point cloud noises is adapted to this task by replacing pre-calibrated 4 cameras with every 4 consecutive frames with known pre-computed poses.  Besides, $5\times5$ median filter based on the depth map and statistical filter~\cite{balta2018fast} based on point cloud are also included in our experiments. We evaluate those models and methods based on the Mean Average Precision where the smeared class is considered positive and the valid point is set as negative. For qualitative comparisons different from others, the predicted results are converted to the point cloud using an intrinsic matrix where smeared points are colored red while the valid points are colored green.

\hspace{-0.4cm}{\bf Implementation Details:} As mentioned, the geometric labels are first built when joining the off-the-shelf semantic segmentation network. A softmax layer is added to adapt to our segmentation task and we use ResNet-34\cite{7780459} as the backbone for UNet~\cite{Unet},  DeepLabV3+~\cite{Deeplabv3plus},  Segformer~\cite{xie2021segformer}. All codes are implemented by Pytorch and all input frames and labels are cropped and resampled to $512\times512$ for computational needs by using nearest-neighbor interpolation to avoid creating artifacts.  Augmentation is performed through random cropping to $128\times128$ with random rotation. We use the mini-batch Adam~\cite{kingma2014adam} optimization algorithm, with a weight decay 1e-7, and run 200 epochs with a batch size 32. The initial learning rate is set at 1e-4 and reduced by 10 times after every 25 epochs with a 100-step cosine annealing schedule~\cite{loshchilov2016sgdr}. We set $\alpha=0.3,\beta=0.7,\epsilon=4mm,\delta=15mm$ in our experiments. The used adjacent reference frame number is $m=4$.
\subsection{Quantitative and Qualitative Results}
\begin{table}[h!]
\centering
\scalebox{0.95}{
\begin{tabular}{cccc}
   \toprule
   Method&Inputs&Features&mAP\\
   \midrule
   Median Filter&$d$&Hand-crafted&0.231\\
   Statistical Filter~\cite{balta2018fast}&$p$&Hand-crafted&0.407\\
 \cline{1-4}\\
    DeepDD~\cite{Self-supervised}
    &$(d,\omega)$&self-annotated&0.103\\
 \cline{1-4}\\
    DeepToF~\cite{10.1145/3130800.3130884}
    &$(d,\omega)$&self-annotated&0.742\\
    DeepLabV3+~\cite{Deeplabv3plus}
    &$(d,\omega)$&self-annotated&0.766\\
    Segformer~\cite{xie2021segformer}
    &$(d,\omega)$&self-annotated&0.729\\
   UNet~\cite{Unet}&$(d,\omega)$&self-annotated&\textbf{0.775}\\
   *UNet~\cite{Unet}&$(d,\omega)$&self-annotated&0.771\\
   \bottomrule
\end{tabular}}
\caption{Results of various methods on our AzureKinect datasets. Each row reports the mean average precision of the smeared points with ground truth. * denotes uniform weighting ($\emph{c}=1$) in \cref{eq:finalloss}.}
\label{table:realsmeared}
\end{table}
\begin{figure*}[h!]
  \centering
    \includegraphics[width=0.77\linewidth]{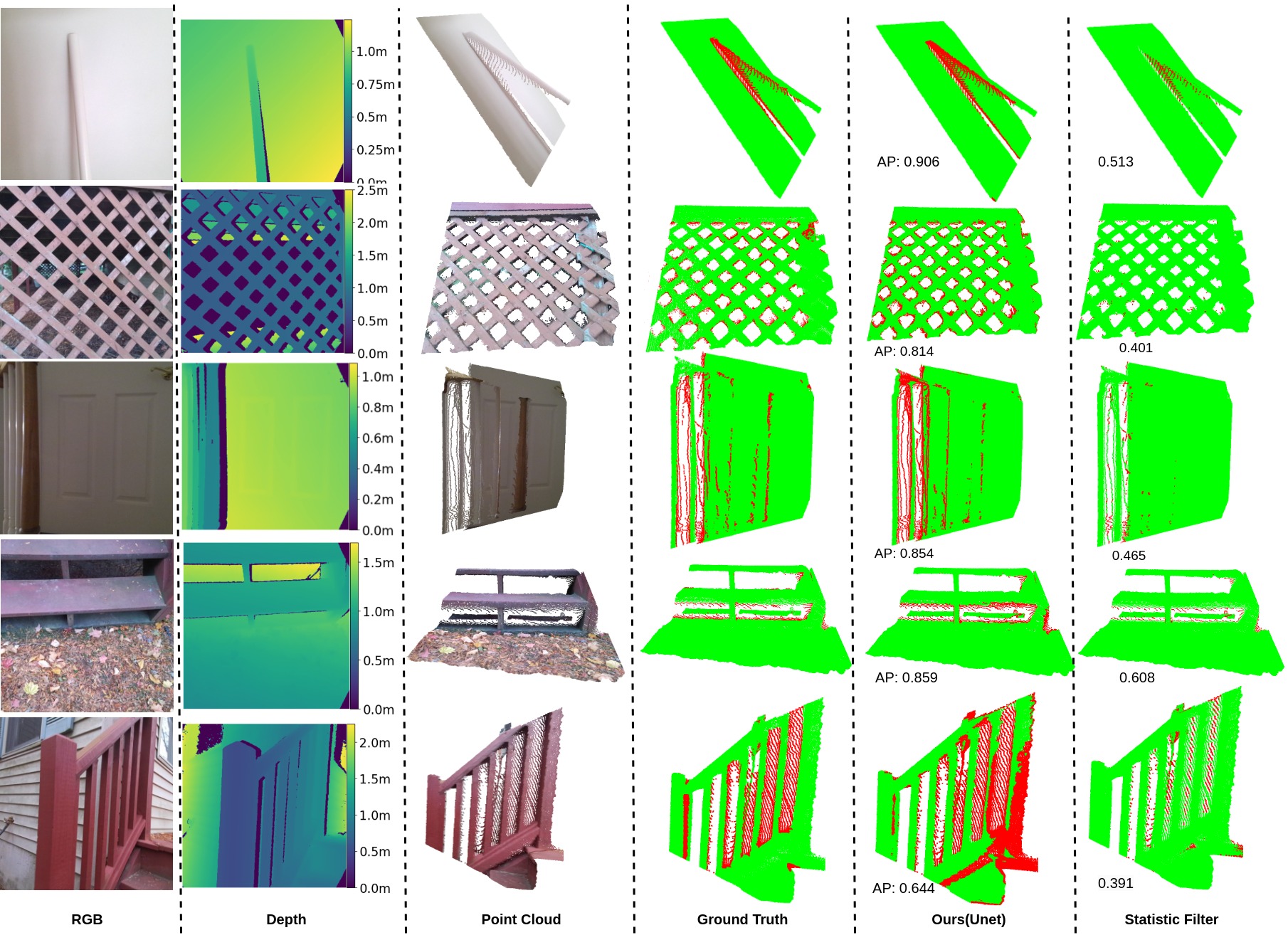}
\caption{Predicted results with AP on our AzureKinect dataset of our self-annotated learning method using UNet and statistical filter. Smeared points are colored red while valid points are colored green. The areas without any point are masked in white.}
\label{fig:plant3d_experiment_2}
\end{figure*}
To obtain pose information, multiview ICP~\cite{segal2009generalized} with five-neighboring point clouds automatically aligns points and determines camera poses. For the DeepDD~\cite{Self-supervised} model which is a regression model compared to our segmentation task, we apply the threshold standard to get evaluation scores by computing abstract differences between the restored depth and raw depth. If the difference is smaller than the Azure Kinect's systematic error threshold ($11mm+0.1\%d$),   then the depth pixel location is predicted valid, otherwise (larger than that threshold) smeared. 
Five cases of the test dataset are shown in \cref{fig:plant3d_experiment_2}, where the self-annotated UNet can detect most of the smeared points than the statistical filter though more valid points are misclassified as the distance increase and it is also challenging for a deep learning remover to detect these smeared points which share the similar structures as valid points, observed in the last row of \cref{fig:plant3d_experiment_2}. We evaluate 11 different depth maps from 11 different scenes, where the model using UNet achieves the highest mAP compared to other methods, see \cref{table:realsmeared}.  Besides, using uniform weighting ($\emph{c}=1$) for multi-view annotation reduces the mAP by $4\%$ than our confidence score design in \cref{equ:confidence_score}. The failure of the self-supervised method DeepDD~\cite{Self-supervised} is also noticed in our experiment, where both the consecutive frames with close viewings, and similar color information among the same observed structures impede this method's effectiveness (please refer to our supplementary materials for more qualitative analysis). 

\subsection{Ablation}
\begin{figure}[h!]
    \centering
    \begin{minipage}{.48\linewidth}
        \centering
        \hspace*{0.2cm}\hspace{-0.730cm}\includegraphics[width=1.18\linewidth]{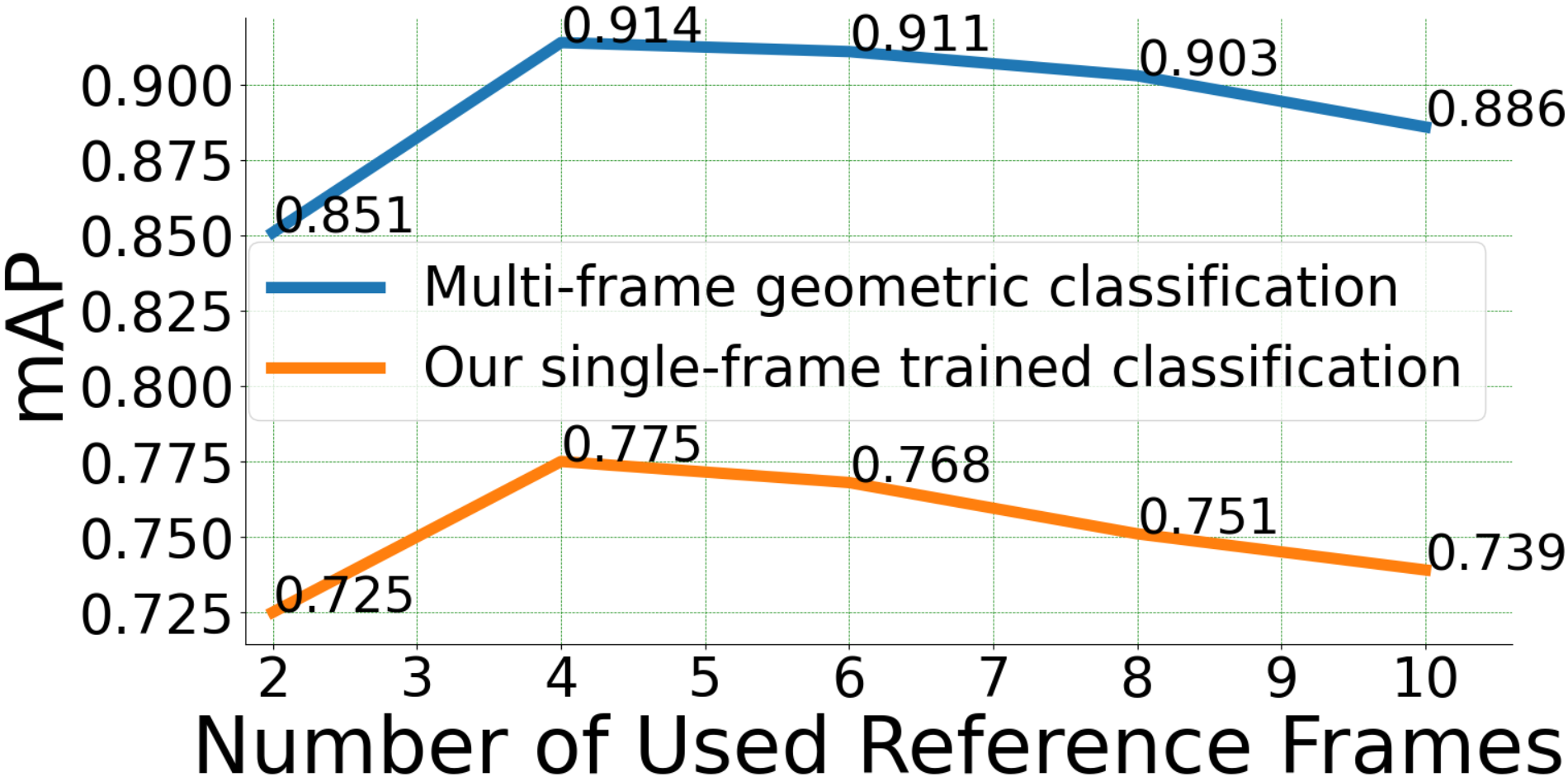}  
        \captionof{figure}{Results of geometric labels generated from different numbers of nearby frames. We report the mean average precision on the AzureKinect data both for multi-frame geometric classification and our single-frame trained classification.}
        \label{fig:many_frames}
    \end{minipage}%
    \hfill
    \hspace{0.18cm}
    \begin{minipage}{.48\linewidth}
          \scalebox{0.65}{
            \raggedright\hspace*{1cm}\begin{tabular}{cccc}
            \toprule
            Depth & Color& $\omega$& AP \\
            \midrule
            \Checkmark &\bf{$-$}&\Checkmark& {\bf 0.775} \\
            \Checkmark &\bf{$-$}&\bf{$-$}&0.670 \\
            \bf{$-$} & \Checkmark &\bf{$-$}&0.567 \\
            \bf{$-$} & \Checkmark &\Checkmark&0.629 \\
            \Checkmark  & \Checkmark &\bf{$-$}&0.613 \\
            \Checkmark & \Checkmark &\Checkmark&0.691 \\
            \bottomrule
            \end{tabular}}
            \captionof{table}{Results of UNet (ours) with different input types. We report the Average Precision (AP) on the AzureKinect data after hyperparameter optimization.}
            \label{tab:ablation1}
    \end{minipage}
\end{figure}

To identify the optimal number of consecutive reference frames required, we repeat experiments with different self-annotated labels for partial points, each derived from different numbers of reference frames. We also generate such labels for the test set to ascertain the accuracy of our geometry annotation.  Both evaluations on multi-frame geometric classification and our single-frame trained classification are concluded as in \cref{fig:many_frames}. Geometry labels for partial points exhibit $12\%-15\%$ \textbf{higher} mAP than UNet for all points, affirming the precision of our self-annotated labels for partial points.
Moreover, using more frames doesn't feed better labels back since the pose estimation is less accurate for long-distance frames and the contradictory information from different frames stands out which further prevents predicted improvements when using more frames. 

To validate our selection for input modality $\phi$, we replace our remover's input with multiple different combinations of color, depth, and normal-view map $\omega$ and evaluate it after 100 training epochs (all convergence guaranteed). For a fair comparison, we conduct a hyperparameter search for each kind of input modality $\phi$ and report results in \cref{tab:ablation1} which show that the $\omega$ map helps detect smeared points both for depth map and color map with a large increase. Besides, indicated by the drop in performance, we think color images contain some invalid information from similar visual features and produce disturbances.

\begin{figure}[h]
  \centering
    \includegraphics[width=1\linewidth]{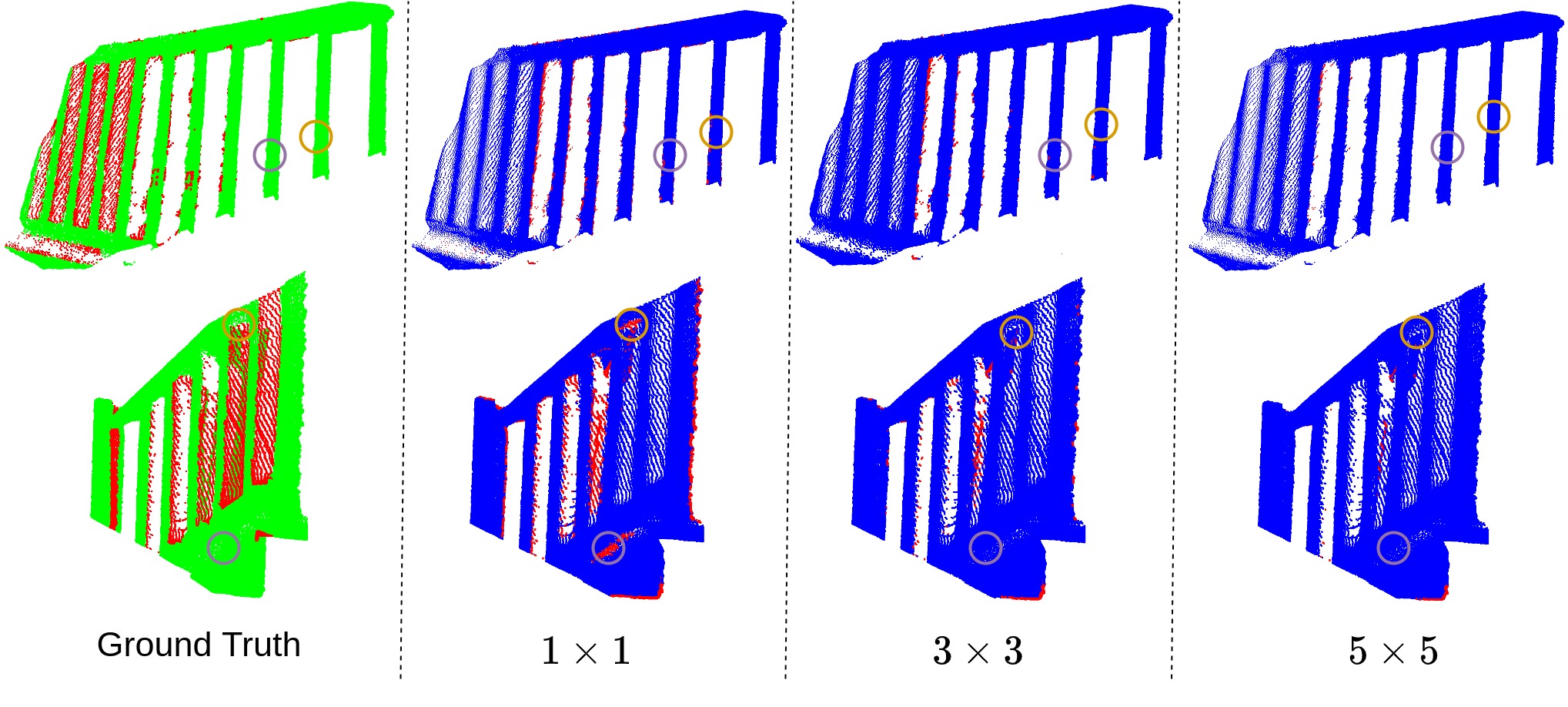} 
\caption{Qualitative comparison among different sliding window sizes for reducing unconfident labels from See-through Empty. The remaining smeared points are colored red with left blue. Misclassifications are reduced and can be seen in small circles.}
\label{fig:see-through_empty}
\end{figure}

To validate our choice for the sliding window size $\varphi=3\times3$ in reducing unconfident self-annotated smeared labels in See-Through Empty, different kernel sizes are applied as shown in \cref{fig:see-through_empty} for the qualitative comparisons. When $\varphi=1\times1$, it is equivalent not to filter any self-annotated smeared points from See-through Empty. Both $3\times3$ and $5\times5$ effectively avoid some misclassifications,  but the sliding window with size $3\times3$ can keep more confident smeared labels than that of $5\times5$. With $\varphi>5\times5$,  too few smeared points are expected to be detected. Therefore, our selection for the sliding window is based on a trade-off assessment of self-annotated label quality and quantity.

\subsection{Application: 3D Reconstruction}
\begin{figure}[h!]
  \centering
    \includegraphics[width=1.0\linewidth]{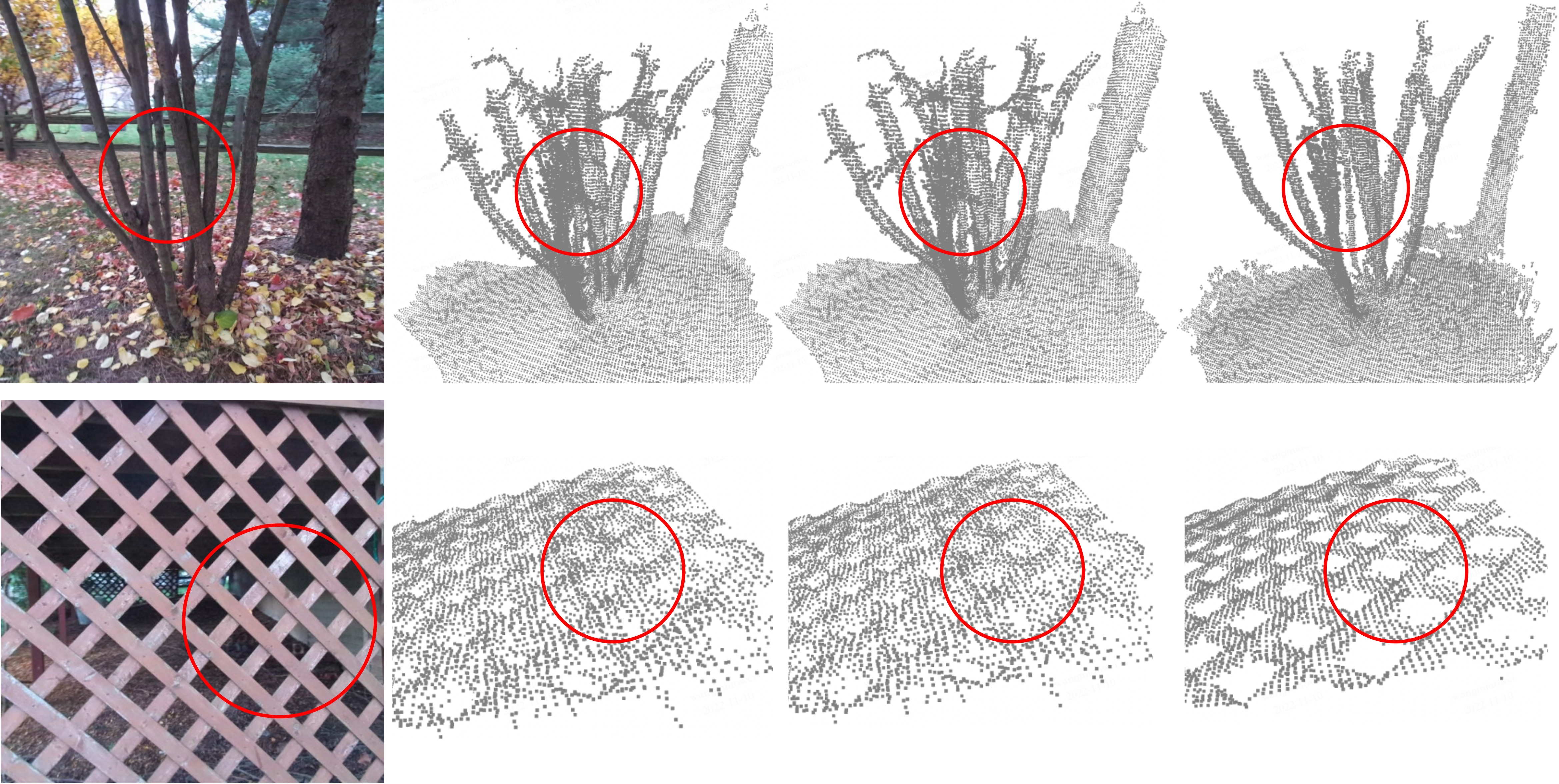}
\caption{Results of multiple frames alignments using the trained network. From the left column to the right, the second column is the aligned point cloud without any filtering; the third column is the aligned point cloud adding an outlier filter; the last column is using our network as a preprocessor for the raw depth map.}
\label{fig:application}
\end{figure}

It is always a major challenge to reconstruct objects with sophisticated fine-grained structures using consumer-level cameras. A related experiment in \cref{fig:application} aligns 15 consecutive frames under the 5HZ work frequency of an Azure Kinect depth sensor and uses down-sampling to make a number of point clouds consistent with three different pre-processes: without any filtering, adding a statistic outliers filter, or using trained UNet model as a preprocessor. Qualitative result in \cref{fig:application} shows that our trained removal better helps align and keep high-fidelity 3D point clouds relieved of smeared points when placed as a preprocessor.
\subsection{Limitations}
Our pipeline still has several limitations.  First, the scenes for training, although not inference, must be static which reduces our data selection especially outside. Second, mechanisms encouraging models to connect and attribute predictions among similar 3D geometry structures need to be further investigated since self-annotated labels are partial and not enough. Finally, incorrect pose estimation due to smeared points can lead to errors. An experiment is performed, where we repeat pose estimation again only using detected valid points (from our initially trained filter), regenerate pseudo-labels, and then retrain our remover from scratch. Results show APs of generated pseudo labels for partial points and predicted scores for all points are raised by 1.5\% and 0.8\% respectively.  

\section{Conclusion}
In this work, we present a new self-annotated architecture to detect smeared points and then remove this harmful artifact from consumer depth sensors.  Visibility-based evidence is automatically gathered from multiple viewpoints of a hand-held sensor to annotate depth pixels as smeared valid or unknown.  These annotations are used to train our smeared point detector with no need for manual supervision.   Being self-annotated avoids the need for costly human annotation while enabling simple data collection and training of widely varied scenes.  As a low-computational network, it can be used as a preprocessor for every single raw frame to improve the quality of 3D reconstruction.
\clearpage

{\small
\bibliographystyle{ieee_fullname}
\bibliography{egbib}
}

\end{document}